%% file: main.tex
\PassOptionsToPackage{table}{xcolor}
\documentclass[11pt,a4paper,logo,nonumbering]{style}
\usepackage{array,booktabs,tabularx,multirow,graphicx}
\usepackage{tikz,pgfplots}
\pgfplotsset{compat=1.18}
\usepgfplotslibrary{groupplots}
\usepackage{tcolorbox}
\tcbuselibrary{breakable}
\usepackage{adjustbox}
\usepackage{placeins}
\definecolor{brandorange}{HTML}{ED6000}
\definecolor{absframe}{HTML}{ED6000}
\newcommand{\lowrate}{1.25\!\times\!10^{-5}}
\newcommand{\highrate}{2.5\!\times\!10^{-5}}
\newcommand{\se}[2]{#1\!\pm\!#2}
\graphicspath{{assets/}}
\hypersetup{pdftitle={Improving JEV Entailment Reranking with Span Readouts and Selective Adaptation},pdfauthor={Haocheng Mei, Yuchuan Tian, Ye Yuan, Enhua Wu, Chao Xu, Allen Chen, Kai Han, Yunhe Wang}}
\makeatletter
\renewcommand{\abscontent}{%
  \begin{tcolorbox}[breakable,colframe=absframe,colback=absframe!5,boxrule=1pt,arc=8pt,
    left=10pt,right=10pt,top=10pt,bottom=10pt]
  \centerline{\fontsize{15pt}{14pt}\selectfont\textbf{Abstract}}\vspace{2ex}
  {\absfont\theabstract}
  \end{tcolorbox}%
}
\renewcommand{\maketitle}{%
  \begingroup\setlength{\parindent}{0pt}
  \vspace*{3pt}
  {\centering\titlefont\@title\par}
  \vskip12pt
  {\centering\@author\par}
  \vskip14pt
  \endgroup
  \thispagestyle{firststyle}\abscontent
}
\makeatother
\fancypagestyle{firststyle}{%
  \fancyhf{}
  \fancyhead[L]{\smash{\raisebox{0.15em}{\includegraphics[height=1.15em]{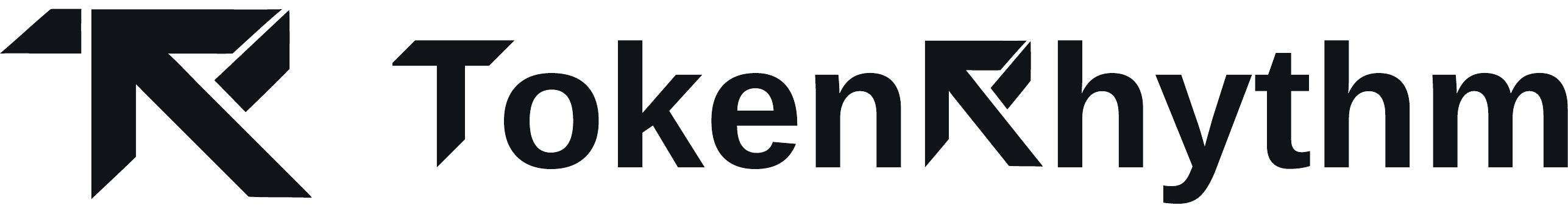}}}}

}
\title{Improving JEV Entailment Reranking with Span Readouts and Selective Adaptation}
\renewcommand\Affilfont{\normalfont\fontsize{10}{12}\selectfont\centering}
\author[1]{Haocheng Mei}
\author[1]{Yuchuan Tian}
\author[2]{Ye Yuan}
\author[3]{Enhua Wu}
\author[1]{Chao Xu}
\author[4]{Allen Chen}
\author[5]{Kai Han}
\author[5]{Yunhe Wang}
\makeatletter
\@namedef{@sep5}{\protect\\\protect\Authfont}
\renewcommand\AB@affilsepx{\protect\\\protect\Affilfont}
\makeatother
\affil[1]{Peking University}
\affil[2]{SUIBE}
\affil[3]{University of Macau}
\affil[4]{Tsinghua University}
\affil[5]{TokenRhythm Technologies}
\date{}
\begin{abstract}
\input{abstract}
\end{abstract}
\begin{document}
\maketitle
\input{opening}
\input{design}

\input{evaluation}

\input{related}
\section{Conclusion}
Fixed-checkpoint comparisons show that direct selection and entailment reranking can favor different option readouts, indicating that design choices for single-request selection need not transfer unchanged to cross-request reranking. Motivated by this observation, we pair mean-span pooling over each NLI option with selective backbone adaptation restricted to attention and DeltaNet projections, jointly adjusting how option states are extracted and how the backbone is updated. Matched comparisons under a tuned one-epoch recipe establish the gains from both choices, while pointer-position, parameter-budget, and head-capacity controls clarify their respective contributions.

On NeoHorse-1-4B, selective adaptation improves MC11 from 50.28\% to 52.49\%, and adding mean-span pooling reaches 53.39\%, with both steps positive in all four fold--seed settings. Together, these changes improve reranking accuracy by about 3.1 pp while reducing trainable parameters by 55.2\%.
\FloatBarrier
\clearpage
\begingroup
\small\setlength{\bibsep}{0.5pt plus 0.25pt}

\input{refs}
\endgroup
\clearpage
\appendix
\input{protocol}

\input{supplement}
\end{document}

%% file: abstract.tex
JEV decision models pair a language model backbone with a lightweight scoring head to select among natural-language options. Direct selection compares candidate answers within one request, whereas entailment reranking evaluates each candidate in a separate natural language inference (NLI) request over fixed relation options. We examine whether the readout and adaptation choices used for direct selection remain effective for cross-request reranking. We combine a mean-span readout with selective backbone adaptation. Mean-span pooling replaces the option-ending pointer with an average of the contextualized states inside each NLI option, without adding head parameters. Selective adaptation retains low-rank updates in attention and gated DeltaNet projections while removing feed-forward adapters. On MC11, which averages accuracy over eleven reranking configurations from seven sources, selective adaptation improves NeoHorse-1-4B accuracy over the all-module baseline while cutting trainable parameters by more than half, and mean-span pooling yields further consistent gains at the same parameter count.
Project page: \url{https://github.com/Mei-TN/Jev-reranking}.

%% file: opening.tex
\section{Introduction}
JEV models turn a natural-language state into a structured decision that software can act on. Their typed interfaces include \texttt{Choice}, which selects among described options; \texttt{Noul}, which answers a binary question; and \texttt{Score}, which assigns probabilities to ordered levels \citep{typesafe,ling}. Open implementations such as NeoHorse-Jev and Kev attach a small bilinear scoring head to an adapted language model backbone \citep{tokenrhythm,kev}, encoding the context and options in one forward pass and using the hidden state at each option's closing delimiter to produce a distribution over the candidates.

These models can answer multiple-choice questions either by comparing all candidate answers within one request (direct selection) or by evaluating each candidate in a separate natural language inference (NLI) request (entailment reranking). Reranking provides a fixed scoring interface for externally supplied answers: their number and wording can vary while the head continues to score entailment, neutral, and contradiction. The question serves as the premise and each candidate answer enters the instruction as a hypothesis; selection then compares entailment probabilities across requests. Because the NLI option text is fixed while its contextualized states vary with the candidate hypothesis, designs suited to comparing answers within one request may be less effective when candidates are scored independently. We therefore combine mean-span pooling over the interior option states with selective backbone adaptation, gathering information from more positions while updating fewer backbone projections.

The two design choices jointly adjust how option representations are extracted and how they are learned. First, a \emph{mean-span readout} replaces the option-ending delimiter pointer with the mean of the contextualized states inside each NLI option. Because the bilinear head is affine in the option states under a shared decision query, pooling computes an exact average of pre-softmax token-level scores and distributes the direct scoring gradient across the span without adding head parameters. Second, \emph{selective backbone adaptation} retains low-rank (LoRA) updates in the attention and gated DeltaNet projections that participate in contextual interaction, removing feed-forward network (FFN) adapters while keeping the pretrained FFN weights in the forward pass. Because the head and adapters shape the option representations together, we evaluate both designs under a common one-epoch, lower-learning-rate recipe that avoids the accuracy loss observed with longer training or higher adapter rates.

Our experiments on NeoHorse-1-4B show that the scoring interface changes the relative performance of option readouts, while reducing adaptation capacity can improve reranking accuracy. With checkpoints and questions held fixed on PIQA and SocialIQA, the pointer leads on average under direct selection, while pooling leads under entailment reranking. On the eleven-configuration MC11 benchmark, the one-epoch, lower-rate recipe establishes an all-module pointer reference at 50.28\% accuracy. Selective adaptation raises this to 52.49\% while reducing trainable parameters by 55.2\% (from 131.2M to 58.8M), and adding mean-span pooling lifts accuracy further to 53.39\% at the same parameter count. Both steps improve MC11 across all four fold--seed settings. Under all-module adaptation, mean-span pooling (51.00\%) also outperforms a pointer at the final content token (49.11\%). The individual design choices also improve PIQA and CommonsenseQA reranking, whereas the combined model loses accuracy on the JevBench and Kev direct-decision suites.

\section{Readout and Adaptation Design}\label{sec:method}
\subsection{Typed requests and candidate reranking}
A request contains a state $x$, an instruction $u$, and named options $O=\{o_1,\ldots,o_K\}$. Each option comprises a label and description between explicit delimiters. The instruction precedes the options; option isolation is disabled, so an option state can attend to the instruction and earlier options. The model returns a distribution over the options.

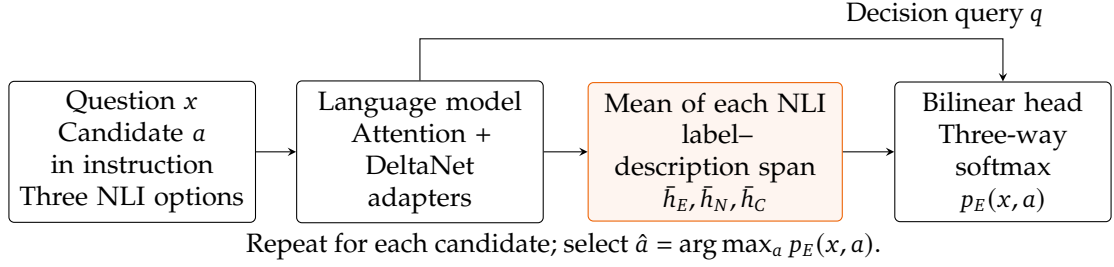
\begin{figure}[!htbp]
\centering
\begin{tikzpicture}[>=stealth, every node/.style={font=\small},
block/.style={draw,rounded corners=2pt,align=center,minimum height=1.55cm}]
\node[block,text width=3.0cm] (input) at (1.6,0)
{Question $x$\\Candidate $a$ in instruction\\Three NLI options};
\node[block,text width=3.0cm] (backbone) at (5.4,0)
{Language model\\Attention + DeltaNet\\adapters};
\node[block,draw=brandorange,fill=brandorange!6,text width=3.1cm] (pool) at (9.3,0)
{Mean of each NLI\\label--description span\\$\bar h_E,\bar h_N,\bar h_C$};
\node[block,text width=2.6cm] (score) at (13.1,0)
{Bilinear head\\Three-way softmax\\$p_E(x,a)$};
\draw[->] (input) -- (backbone);
\draw[->] (backbone) -- (pool);
\draw[->] (pool) -- (score);
\draw[->] (backbone.north) -- +(0,0.55) -| node[pos=0.45,above]{Decision query $q$} (score.north);
\node[align=center] at (7.3,-1.25)
{Repeat for each candidate; select $\hat a=\arg\max_a p_E(x,a)$.};
\end{tikzpicture}
\caption{The final reranker. The candidate answer enters the instruction; pooling reads the three NLI option spans. The shared head scores those spans against the decision query, and the outer comparison selects an answer using its entailment probability.}
\label{fig:reranker}
\end{figure}

Direct selection chooses the highest-probability option in one request. For entailment reranking, the state supplies the premise, and each candidate answer $a$ is rendered as a hypothesis in the instruction of a separate request. Its three options are entailment, neutral, and contradiction. We select
\begin{equation}
\begin{aligned}
\hat a&=\arg\max_a p_E(x,a),\\
p_E&=\frac{e^{z_E}}{e^{z_E}+e^{z_N}+e^{z_C}},
\end{aligned}
\label{eq:rerank}
\end{equation}
where $z_E,z_N,z_C$ are the three option logits. The head reads the NLI labels and descriptions. Although this text is shared across candidates, its hidden states depend on the preceding question in the state and the candidate hypothesis in the instruction. Each request therefore supplies a distinct contextual representation of the three relations (Figure~\ref{fig:reranker}).

Reranking uses the full probability of entailment, even when several candidates receive entailment as their highest-scoring label. Equivalently,
\begin{equation}
p_E=\sigma\!\left(z_E-\log(e^{z_N}+e^{z_C})\right),
\label{eq:margin}
\end{equation}
where $\sigma$ is the logistic function. Candidates are ordered by an entailment margin against the combined alternatives. A change to the three scores can alter candidate ordering without changing the winning NLI label within individual requests.

\subsection{Option-ending and mean-span readouts}
Let $h_d\in\mathbb R^d$ be the hidden state at the final decision token, which follows all options, and $h_i\in\mathbb R^d$ the state at the delimiter closing option $i$. Following the released NeoHorse-Jev implementation \citep{tokenrhythm}, the pointer uses shared affine query and key projections:
\begin{align}
q &= W_qh_d+b_q,\quad k_i = W_kh_i+b_k,\\
s_i &= \frac{q^\top k_i}{\sqrt{d_{\mathrm{head}}}},\quad
p_i = \frac{e^{s_i}}{\sum_j e^{s_j}},
\end{align}
with $d_{\mathrm{head}}=256$. Training minimizes cross-entropy against the target option distribution, averaging questions within each record.

Mean-span pooling replaces $h_i$ with
\begin{equation}
\bar h_i=\frac{1}{|S_i|}\sum_{j\in S_i}h_j,
\end{equation}
where $S_i$ contains the option's interior label and description tokens, excluding both delimiters. The last-content control instead takes the final state in $S_i$. Both rules fall back to the closing delimiter for empty content. All three readouts use the same query and $512(d+1)$ head parameters, and each model is trained and evaluated with its own readout.

The affine projection makes the pooled score an exact average of the contextual token scores. For a nonempty span,
\begin{align}
s_i^{\mathrm{pool}}
&=\frac{q^\top(W_k\bar h_i+b_k)}{\sqrt{d_{\mathrm{head}}}}\nonumber\\
&=\frac{1}{|S_i|}\sum_{j\in S_i}
\frac{q^\top(W_kh_j+b_k)}{\sqrt{d_{\mathrm{head}}}}.
\label{eq:pool-score}
\end{align}
The averaging takes place before the option softmax, with span-length normalization giving each option one mean score. All token contributions use the same key projection and decision query.

Pooling also changes the direct training signal. Holding the query fixed, each interior state receives the score gradient
\begin{equation}
\frac{\partial s_i^{\mathrm{pool}}}{\partial h_j}
=\frac{W_k^\top q}{|S_i|\sqrt{d_{\mathrm{head}}}},\qquad j\in S_i.
\end{equation}
The pointer supplies this direct path through one selected state; pooling supplies it through every interior state. Contextualization provides additional paths through the backbone in both cases.

\subsection{Selective adaptation}
We jointly train the decision head and LoRA adapters \citep{hu}, comparing two scopes: all-module adaptation covers attention, gated DeltaNet, and feed-forward network (FFN) projections, whereas selective adaptation (no-FFN) retains adapters only in attention and DeltaNet. Both configurations keep every pretrained FFN layer in the forward computation. Crossing these scopes with the pointer and mean-span readouts lets us compare the readout choices under both forms of adaptation while keeping head size and adapter rank fixed.

At inference, pooling averages hidden states already produced by the backbone before applying the key projection (Equation~\ref{eq:pool-score}). Each option therefore requires one projected key, and reranking continues to use one backbone request per answer. Selective adaptation reduces the parameters requiring gradients and optimizer states during training, while the pretrained backbone remains intact.

\section{Experimental Setup}\label{sec:setup}
\paragraph{Models and training.}
Our main experiments use NeoHorse-1-4B \citep{neo,neocard}, a post-trained, text-only repackaging of Qwen3.5-4B \citep{qwen35}. Its backbone combines attention and gated DeltaNet layers \citep{deltanet}. The reference recipe trains the decision head and rank-64 adapters jointly for one epoch with a cosine variant of the 1cycle policy \citep{smith}, using a peak learning rate of $\lowrate$ for both parameter groups. The learning-rate diagnostics also test $\highrate$, and all comparisons evaluate the final weights of each run. A separate pointer--pooling comparison on Qwen3-4B \citep{qwen3} examines the readout change with another backbone.

The training mixture combines web decisions and policy tasks with sentiment, topic, intent, and entailment classification. Public-source labels and model-generated reference answers are converted to a shared request format while retaining each question's native options and target distribution. MC11 reuses the scorer learned from this mixture: each candidate answer becomes a separate NLI request, and its entailment probability determines its rank among the candidates.

To keep related records together, we assign whole groups to splits before forming eight training pools, which contain 17,661 records and 24,001 questions. Each fold trains on seven pools; crossing two folds with two random seeds gives four matched settings. Within a setting, each configuration is trained afresh with compatible initial tensors, data order, and augmentation streams matched across the comparison. The learning-rate, scope, and budget studies reuse the same lower-rate, one-epoch reference pointers, giving their contrasts a common reference within each setting. Appendix~\ref{app:protocol} gives the grouping rules, augmentation procedure, and complete training configuration.

\paragraph{Evaluation.}
MC11 averages candidate-reranking accuracy across ARC-Easy and ARC-Challenge \citep{arc}; four- and ten-option GSM8K \citep{gsm8k}; GPQA and MMLU with and without demonstrations \citep{gpqa,mmlu}; HellaSwag \citep{hellaswag}; WinoGrande \citep{winogrande}; and synthetic chess. Each configuration receives equal weight and scores its candidates through Equation~\ref{eq:rerank}. The paired configurations of GSM8K, GPQA, and MMLU reuse their respective question populations, changing the candidate count or demonstration context.

Candidate construction preserves the supplied answers for multiple-choice tasks. GSM8K pairs the numerical answer with three or nine generated distractors; synthetic chess pairs one legal move with three illegal moves. WinoGrande fills the sentence gap, and HellaSwag uses the candidate ending as the NLI hypothesis.

The additional reranking evaluations use PIQA and CommonsenseQA \citep{piqa,csqa}; fixed-checkpoint PIQA and SocialIQA tests compare both interfaces on the same questions \citep{socialiqa}. For JevBench, we report family-macro accuracy on its frozen 231-question public subset, equally averaging 18 families \citep{jevbench}. Our Kev endpoint equally averages six development/test selections from its decision suite and two transfer suites \citep{kev}.

\paragraph{Reporting.}
We report accuracy in percent and paired changes in percentage points (pp). For each contrast, we first calculate the difference within each of the four settings and then report its mean and descriptive SE; SEs for absolute accuracy are calculated directly from the four model scores. These SEs summarize variation across settings that share training pools and evaluation questions. Appendix~\ref{app:robust} examines sensitivity to source weighting, omission of individual sources, and repetition of the reference run.

\section{Readout Behavior Across Decision Interfaces}\label{sec:interface}
We compare pointer and pooling models under both request interfaces to test whether the relative performance of the readouts changes with how answers are scored. Each checkpoint is held fixed across interfaces, and both interfaces use the same PIQA and SocialIQA questions. PIQA contains 1,838 validation questions and uses eight higher-rate checkpoints. SocialIQA contains 1,954 development questions and uses eight checkpoints at each of two training budgets.

\begin{table}[!htbp]
\centering\small
\begin{tabular}{lrrrrrr}
\toprule
 & \multicolumn{2}{c}{Direct accuracy (\%)} & \multicolumn{2}{c}{NLI accuracy (\%)} & \multicolumn{2}{c}{Pooling minus pointer (pp)}\\
Population and training & Pointer & Pooling & Pointer & Pooling & Direct & NLI\\
\midrule
PIQA, higher-rate models &79.33&79.20&65.15&73.22&$\se{-0.12}{1.41}$&$\se{+8.07}{3.76}$\\
SocialIQA, one epoch &73.76&71.43&61.12&61.40&$\se{-2.33}{0.64}$&$\se{+0.28}{0.72}$\\
SocialIQA, two epochs &73.67&71.81&56.82&58.84&$\se{-1.86}{0.23}$&$\se{+2.02}{1.20}$\\
\bottomrule
\end{tabular}
\caption{The request interface changes the readout comparison on the same questions and fixed checkpoints. Accuracies are four-setting means; differences include descriptive SEs. Neither source enters MC11.}
\label{tab:interface}
\end{table}

On PIQA, the readouts are nearly tied under direct selection, whereas pooling leads the pointer under reranking: the relative difference moves from $-0.12$ to $+8.07$ pp (Table~\ref{tab:interface}). SocialIQA shows the same direction of interface interaction at both budgets, with a positive interaction in every setting. Direct selection has higher mean absolute accuracy on both sources; within reranking, pooling becomes the stronger readout on average.

The readout comparison also depends on the NLI option wording. Across 18 combinations of three descriptions and six label orders on PIQA, pooling's mean advantage is $+3.18\pm3.27$~pp; one description set instead favors the pointer by 1.80~pp after averaging over orders (Appendix~\ref{app:additional}).

%% file: design.tex
\section{Reranking Accuracy and Parameter Efficiency}\label{sec:design}
The core comparisons use the same one-epoch recipe, with peak learning rates of $\lowrate$ for both the head and adapters and adapter rank 64. We first examine the combined design, then use controls on adapter allocation, readout, and training to explain the choices.

\subsection{Combining Selective Adaptation and Span Pooling}
Combining the two choices gives the highest MC11 accuracy among the four matched configurations, with less than half the trainable parameters of the all-module models (Table~\ref{tab:factorial}). Starting from the all-module pointer at 50.28\%, removing FFN adapters raises accuracy to 52.49\%, a paired gain of $2.21\pm0.66$ pp. Mean-span pooling then raises it to 53.39\% at the same parameter count, adding $0.91\pm0.35$ pp. Both steps improve MC11 in every fold--seed setting. Removing FFN adapters reduces trainable parameters from 131.17M to 58.82M, a 55.2\% reduction (Table~\ref{tab:capacity}).

\begin{table}[!htbp]
\centering\normalsize
\captionsetup{width=0.88\linewidth}
\begin{tabular*}{0.84\linewidth}{@{\extracolsep{\fill}}lrr@{}}
\toprule
 & \multicolumn{2}{c}{Adapter scope}\\
Readout & All modules & No-FFN\\
\midrule
Pointer & $\se{50.28}{1.02}$ & $\se{52.49}{0.46}$\\
Mean-span & $\se{51.00}{0.21}$ & $\mathbf{53.39}\pm0.24$\\
\midrule
Pool $-$ pointer (pp) & $\se{+0.73}{1.07}$ & $\se{+0.91}{0.35}$\\
Trainable parameters (M) & 131.17 & 58.82\\
\bottomrule
\end{tabular*}
\caption{MC11 accuracy (\%) and trainable parameter count for the core readout-by-scope comparison. All four models use one epoch, lower head and adapter rates, and rank 64. Accuracies and paired changes are four-setting means with descriptive SEs. Both readouts have the same parameter count within each scope.}
\label{tab:factorial}
\end{table}

\begin{table}[!htbp]
\centering\normalsize
\captionsetup{width=0.80\linewidth}
\begin{tabular*}{0.76\linewidth}{@{\extracolsep{\fill}}lrr@{}}
\toprule
Parameter group & All modules & No-FFN\\
\midrule
Attention adapters & 12.58 & 12.58\\
DeltaNet adapters & 44.92 & 44.92\\
FFN adapters & 72.35 & 0\\
Bilinear head & 1.31 & 1.31\\
\midrule
Total trainable & 131.17 & 58.82\\
\bottomrule
\end{tabular*}
\caption{Trainable parameters in millions at rank 64, rounded independently. The pretrained backbone weights are frozen. Both readouts use the same bilinear head.}
\label{tab:capacity}
\end{table}

The final combination gains about 3.1 pp over the reference pointer. Its readout-by-scope interaction is small ($+0.18\pm0.74$ pp), consistent with approximately additive gains from span aggregation and scope restriction.

\subsection{Adapter Scope and Parameter Budget}
FFN adapters account for 72.35M trainable parameters, more than half of the all-module adapter budget. Removing them improves the pointer by $2.21\pm0.66$ pp and the pooling model by $2.39\pm0.35$ pp, showing that the scope change helps with either readout.

A similar parameter budget can instead be distributed across all projections at a lower rank. The all-module rank-28 pointer uses 58.12M trainable parameters and reaches 52.40\%, close to the no-FFN rank-64 pointer's 52.49\%. Both allocations improve over the larger all-module reference. The no-FFN allocation reduces the budget relative to that rank-64 all-module reference while retaining rank 64 in attention and DeltaNet, and combines with pooling to give the strongest configuration in Table~\ref{tab:factorial}.

\subsection{Option Aggregation and Scoring Head}
To examine the role of aggregation, we compare mean-span pooling with a different single-token position and a larger scoring head. The last-content control reads the final token inside each option, retaining the same bilinear head and all-module training recipe. It reaches 49.11\% MC11, below both the option-ending pointer and mean-span pooling at 51.00\%. Pooling exceeds this control by $1.89\pm0.97$ pp, supporting aggregation across the span over either tested pointer position. The largest gains over last-content occur on HellaSwag (8.10 pp) and the two ARC tasks; Appendix~\ref{app:evidence} gives the complete task profile.

\paragraph{Keeping the head simple.}\label{sec:residual}
Adding a two-layer residual correction to the all-module pointer increases its trainable parameter count by about 0.13M and yields 50.33\% MC11, close to the reference pointer and its repeated run. Under this recipe, the residual branch provides little additional benefit, so the final design retains the bilinear head and changes the option representation supplied to it. Appendix~\ref{app:residual} examines the residual branch through inference-removal and training controls.

\subsection{Training Recipe and Reference Quality}
Because the head and adapters are trained jointly, the reference model's quality depends on how strongly the backbone is updated. We vary the head and adapter learning rates independently and compare one- and two-epoch schedules, keeping the all-module, rank-64 architecture fixed.

\begin{table}[!htbp]
\centering\normalsize
\captionsetup{width=0.80\linewidth}
\begin{tabular*}{0.76\linewidth}{@{\extracolsep{\fill}}lrr@{}}
\toprule
 & \multicolumn{2}{c}{Adapter peak rate}\\
Head peak rate & Lower & Higher\\
\midrule
Lower & $\se{50.28}{1.02}$ & $\se{46.81}{2.37}$\\
Higher & $\se{49.66}{1.24}$ & $\se{47.46}{1.35}$\\
\midrule
 & \multicolumn{2}{c}{Training budget, both rates lower}\\
Readout & One epoch & Two epochs\\
\midrule
Pointer & $\se{50.28}{1.02}$ & $\se{43.96}{0.91}$\\
Mean-span & $\se{51.00}{0.21}$ & $\se{48.57}{1.16}$\\
\bottomrule
\end{tabular*}
\caption{Training-recipe diagnostics on MC11 (\%, mean $\pm$ descriptive SE). The upper panel contains one-epoch pointers; the lower panel compares fresh runs with budget-specific schedule horizons. Lower/higher rates are $\lowrate$/$\highrate$.}
\label{tab:recipe}
\end{table}

\paragraph{Learning rates.}
Lowering the adapter rate improves MC11 at both head rates, with an average gain of $2.83\pm0.79$ pp (Table~\ref{tab:recipe}). Changing the head rate has little average effect ($+0.02\pm0.43$ pp). Among these rate settings, the pointer performs best with the lower rate for both parameter groups.

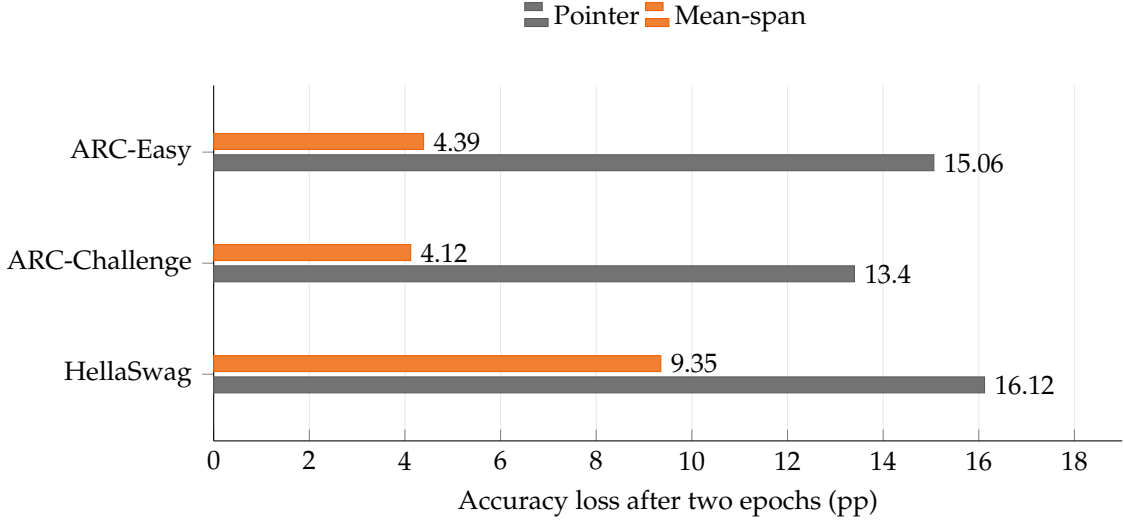
\begin{figure}[!htbp]
\centering
\begin{tikzpicture}
\begin{axis}[
 scale only axis,width=0.75\linewidth,height=4.7cm,
 xbar,bar width=6pt,
 xmin=0,xmax=19,
 symbolic y coords={HellaSwag,ARC-Challenge,ARC-Easy},
 ytick=data,
 tick label style={font=\small},
 label style={font=\small},
 xlabel={Accuracy loss after two epochs (pp)},
 enlarge y limits=0.3,
 legend style={font=\small,at={(0.5,1.13)},anchor=south,legend columns=2,draw=none},
 nodes near coords,point meta=x,
 nodes near coords style={font=\small,/pgf/number format/fixed,/pgf/number format/precision=2},
 xmajorgrids=true,grid style={gray!18},
 axis lines*=left,
]
\addplot[fill=black!55,draw=black!65] coordinates {(16.12,HellaSwag) (13.40,ARC-Challenge) (15.06,ARC-Easy)};
\addplot[fill=brandorange!80,draw=brandorange] coordinates {(9.35,HellaSwag) (4.12,ARC-Challenge) (4.39,ARC-Easy)};
\legend{Pointer,Mean-span}
\end{axis}
\end{tikzpicture}
\caption{Accuracy loss from one to two epochs under the lower-rate all-module recipe. Both readouts favor one epoch on ARC and HellaSwag.}
\label{fig:budget}
\end{figure}

\paragraph{Training budget.}
Both readouts perform better after one epoch in every setting, with the largest losses from longer training on ARC and HellaSwag (Figure~\ref{fig:budget}). The two-epoch pointer falls to 43.96\%, while pooling falls to 48.57\%. Pooling's relative advantage therefore grows from 0.73 to 4.61 pp as both models become less accurate. The one-epoch, lower-rate recipe supplies the strongest of these all-module pointer references at 50.28\% for evaluating the two designs.

\paragraph{Update schedule.}
At the higher learning rate, training only the head for the first 30\% of updates before enabling the adapters yields 53.54\% MC11 with a no-FFN pointer (Appendix~\ref{app:schedules}). The core readout-by-scope comparison keeps joint training and the lower rates fixed, so each design change can be assessed under the same recipe. Combining the head-first schedule with mean-span pooling remains untested.

\subsection{Task-Level Sources of Improvement}
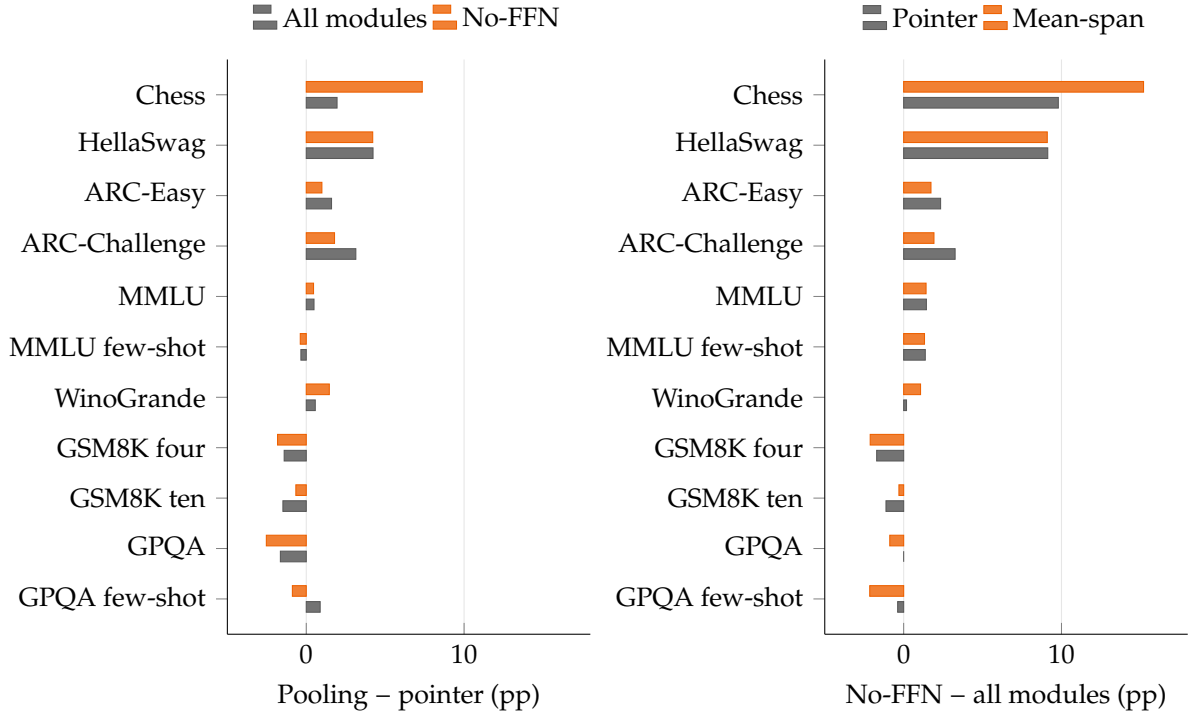
\begin{figure}[!htbp]
\centering
\begin{tikzpicture}
\begin{groupplot}[
group style={group size=2 by 1,horizontal sep=3.1cm},
scale only axis,width=4.8cm,height=7.6cm,
xbar,/pgf/bar width=4pt,
xmin=-5,xmax=18,
symbolic y coords={GPQA few-shot,GPQA,GSM8K ten,GSM8K four,WinoGrande,MMLU few-shot,MMLU,ARC-Challenge,ARC-Easy,HellaSwag,Chess},
ytick=data,yticklabel style={font=\small},
xticklabel style={font=\small},label style={font=\small},
xlabel={Accuracy change (pp)},
enlarge y limits=0.07,
title style={font=\small},
legend style={font=\small,at={(0.5,1.03)},anchor=south,legend columns=2,draw=none},
xmajorgrids=true,grid style={gray!20},axis lines*=left,
]
\nextgroupplot[xlabel={Pooling $-$ pointer (pp)}]
\addplot[fill=black!55,draw=black!65] coordinates {
(0.88,GPQA few-shot) (-1.64,GPQA) (-1.48,GSM8K ten) (-1.40,GSM8K four)
(0.57,WinoGrande) (-0.34,MMLU few-shot) (0.49,MMLU)
(3.14,ARC-Challenge) (1.60,ARC-Easy) (4.23,HellaSwag) (1.95,Chess)};
\addplot[fill=brandorange!80,draw=brandorange] coordinates {
(-0.88,GPQA few-shot) (-2.53,GPQA) (-0.66,GSM8K ten) (-1.82,GSM8K four)
(1.46,WinoGrande) (-0.38,MMLU few-shot) (0.46,MMLU)
(1.79,ARC-Challenge) (0.99,ARC-Easy) (4.21,HellaSwag) (7.35,Chess)};
\legend{All modules,No-FFN}
\nextgroupplot[xlabel={No-FFN $-$ all modules (pp)}]
\addplot[fill=black!55,draw=black!65] coordinates {
(-0.38,GPQA few-shot) (0.00,GPQA) (-1.12,GSM8K ten) (-1.71,GSM8K four)
(0.18,WinoGrande) (1.37,MMLU few-shot) (1.44,MMLU)
(3.26,ARC-Challenge) (2.34,ARC-Easy) (9.13,HellaSwag) (9.80,Chess)};
\addplot[fill=brandorange!80,draw=brandorange] coordinates {
(-2.15,GPQA few-shot) (-0.88,GPQA) (-0.30,GSM8K ten) (-2.12,GSM8K four)
(1.07,WinoGrande) (1.32,MMLU few-shot) (1.42,MMLU)
(1.92,ARC-Challenge) (1.73,ARC-Easy) (9.10,HellaSwag) (15.20,Chess)};
\legend{Pointer,Mean-span}
\end{groupplot}
\end{tikzpicture}
\caption{Task-level effects of the two design choices. Left: mean-span pooling minus the pointer under each adapter scope. Right: removing FFN adapters for each readout. Bars are means over four paired settings under the one-epoch, lower-rate recipe. All configurations retain their original weight in MC11.}
\label{fig:task-gains}
\end{figure}

HellaSwag and synthetic chess account for most of the net MC11 gain (Figure~\ref{fig:task-gains}). On HellaSwag, pooling adds about 4.2 pp under either adapter scope, while removing FFN adapters adds about 9.1 pp with either readout, giving the final model an advantage of about 13.3 pp. On chess, the pooling gain grows from 1.95 pp under all-module adaptation to 7.35 pp without FFN adapters.

The final model also improves both ARC configurations, both MMLU configurations, and WinoGrande, alongside declines on GSM8K and GPQA. Appendix~\ref{app:evidence} gives the component contrasts for every task. The mean pooling and scope gains remain positive under both weighting schemes when any one source is omitted (Appendix~\ref{app:robust}).

%% file: evaluation.tex
\section{Additional Reranking Sources and Backbone Comparison}\label{sec:validation}
\subsection{PIQA and CommonsenseQA}
We test the individual design choices beyond MC11 on PIQA and CommonsenseQA. PIQA asks the model to select between two solutions to a physical goal, whereas CommonsenseQA offers five candidate answers. We apply the same three-option NLI scorer to their 1,838 and 1,221 questions, respectively, using separate matched model groups to compare learning rates, readouts, and adapter scope (Table~\ref{tab:extra}).

\begin{table}[!htbp]
\centering\small
\begin{tabular}{llrr}
\toprule
Factor & Paired comparison & PIQA (pp) & CommonsenseQA (pp)\\
\midrule
Learning rate & Lower-rate pointer $-$ higher-rate pointer & $\se{+12.17}{2.49}$ & $\se{+11.65}{1.57}$\\
Readout & Pooling $-$ pointer, both lower-rate & $\se{+1.26}{0.85}$ & $\se{+4.01}{1.83}$\\
Adapter scope & No-FFN pointer $-$ all-module pointer & $\se{+5.21}{1.15}$ & $\se{+9.56}{2.81}$\\
\midrule
\multicolumn{2}{l}{Scope comparison accuracy: all-module pointer} &75.04\%&54.24\%\\
\multicolumn{2}{l}{Scope comparison accuracy: no-FFN pointer} &80.25\%&63.80\%\\
\bottomrule
\end{tabular}
\caption{Additional-source support for individual design choices. Changes are paired four-setting means with descriptive SEs. The learning-rate comparison changes head and adapter rates together; the readout comparison uses the same model group. Adapter scope is tested separately with lower-rate, one-epoch pointers and matched initializations. These groups are separate from the core readout-by-scope models in Table~\ref{tab:factorial}; the final no-FFN pooling model has not been evaluated on these sources.}
\label{tab:extra}
\end{table}

Lowering both learning rates strengthens the pointer on both sources, and pooling improves these lower-rate models by a further 1.26 pp on PIQA and 4.01 pp on CommonsenseQA. The separate scope comparison follows the same direction: omitting FFN adapters improves both sources in every setting, reaching 80.25\% and 63.80\%, respectively.

\subsection{Readout on Qwen3-4B}
The Qwen3-4B comparison tests the readout change under lower-rate, all-module adaptation. Pooling improves HellaSwag by 3.91 pp, close to the 4.23 pp gain on NeoHorse, and also improves the ARC configurations. Declines elsewhere, including a 4.28 pp loss on four-option GSM8K, outweigh these gains: overall MC11 accuracy is 51.70\% for pooling and 52.17\% for the pointer. The readout change thus retains gains on particular tasks without improving aggregate accuracy on this backbone. Appendix~\ref{app:evidence} reports the complete task comparison.

\section{Trade-offs on Direct Selection}\label{sec:direct}
We also test the same core checkpoints under direct selection to see how the reranking choices affect their original use (Table~\ref{tab:scope}). The no-FFN pointer raises MC11 while maintaining JevBench accuracy, at 77.52\% versus 77.11\% for the all-module pointer. Adding pooling improves MC11 further, with JevBench and Kev accuracies of 73.79\% and 78.12\%. These are 3.32 and 3.25 pp below the all-module pointer.

Across both adapter scopes, the pointer gives higher direct-selection accuracy than pooling on these suites. Kev's declines include decision-test as well as the transfer subsets; Appendix~\ref{app:evidence} gives the six-subset profile. Together with the fixed-checkpoint interface comparisons in Section~\ref{sec:interface}, these results support choosing the readout with the intended scoring task in mind.

\begin{table}[!htbp]
\centering\normalsize
\captionsetup{width=0.80\linewidth}
\setlength{\tabcolsep}{3.5pt}
\begin{tabular*}{0.76\linewidth}{@{\extracolsep{\fill}}lrr@{}}
\toprule
Core model & JevBench (\%) & Kev (\%)\\
\midrule
All-module pointer & $\se{77.11}{0.84}$ & $\se{81.37}{0.59}$\\
All-module pooling & $\se{75.37}{0.96}$ & $\se{79.91}{0.35}$\\
No-FFN pointer & $\se{77.52}{0.98}$ & $\se{79.78}{0.35}$\\
No-FFN pooling & $\se{73.79}{1.18}$ & $\se{78.12}{0.26}$\\
\bottomrule
\end{tabular*}
\caption{Direct-decision results for the core readout-by-scope comparison. Entries are four-setting means with descriptive SEs. MC11 results for these same models appear in Table~\ref{tab:factorial}.}
\label{tab:scope}
\end{table}

%% file: related.tex
\section{Related Work}
\label{sec:related}

Studies of specialized JEV decision models report uneven performance across knowledge, reasoning, and multilingual tasks \citep{li}. Interventions on labels, definitions, and context also change their behavior without changing model weights \citep{azizi,xu}. LLM-as-Jev examines joint suffix scoring without training and fine-tuning with KL anchors \citep{liwagle}. Related work on prompted language models shows that multiple-choice scores depend on the scoring interface and option identifiers \citep{robinson,zheng}. We focus on improving the trained JEV model when it is used to compare candidate answers through separate entailment requests.

Pointer and biaffine architectures score contextual representations with shared projections \citep{vinyals,dozat}. Research on sentence and span representations examines which token states to aggregate \citep{reimers,toshniwal}; recent work extends these choices to language models used as text encoders \citep{wang,behnamghader}. \citet{tang} compare pooling and attention while holding the embedding model's data and backbone fixed. Our comparisons examine option-span readouts trained jointly with adapters, including whether pooling remains useful under a smaller adapter scope.

LoRA restricts adaptation through low-rank updates to selected projections \citep{hu}. Layer selection under distribution shift \citep{lee}, forgetting during adaptation \citep{biderman}, and learning rates across adapter factors and ranks \citep{hayou,chen} provide reasons to examine placement, capacity, and optimization together. Gradual unfreezing \citep{howard}, head fitting before fine-tuning \citep{kumar}, and parameter-efficient head initialization followed by full fine-tuning \citep{yanghead} offer other ways to control adaptation. Fine-tuning stability also depends on initialization, data order, and optimization \citep{dodge,mosbach}. Our paired comparisons bring readout design and adaptation scope together in a typed decision model used for cross-request entailment reranking.

%% file: protocol.tex
\section{Protocol and Reproducibility}
\label{app:protocol}

\subsection{MC11 populations and scoring}
\label{app:mc11_protocol}

MC11 averages the eleven configuration accuracies listed in Table~\ref{app:mc11_inventory}, each computed by ranking candidates with Equation~\ref{eq:rerank}. These configurations contain 46,475 question occurrences per model: GSM8K, GPQA, and MMLU repeat their question populations under two formats, whereas ARC-Easy and ARC-Challenge use different populations from the same source. An occurrence identifies a question within a configuration, so a repeated question contributes once in each configuration. Appendix~\ref{app:robust} examines how the results change when sources receive equal weight or are omitted individually.

\begin{table}[!htbp]
\centering
\small
\begin{tabularx}{\textwidth}{lrlX}
\toprule
Configuration & Questions & Split & External candidates and demonstrations \\
\midrule
ARC-Easy & 2,376 & Test & Choices supplied with each source question. \\
ARC-Challenge & 1,172 & Test & Choices supplied with each source question. \\
WinoGrande & 1,267 & Validation & Supplied alternatives, inserted into the incomplete sentence. \\
GSM8K, four options & 1,319 & Test & Correct numerical answer and three generated distractors. \\
GSM8K, ten options & 1,319 & Test & Correct numerical answer and nine generated distractors. \\
GPQA & 198 & Diamond & Supplied answers, with the correct answer first in loaded order. \\
GPQA, few-shot & 198 & Diamond & Same answers; five solved examples from other Diamond questions. \\
Chess & 500 & Generated & One legal move and three illegal moves. \\
HellaSwag & 10,042 & Validation & Supplied candidate endings. \\
MMLU & 14,042 & Test & Supplied answer choices. \\
MMLU, few-shot & 14,042 & Test & Same choices; five examples from the subject's development split. \\
\bottomrule
\end{tabularx}
\caption{MC11 populations, candidate construction, and demonstration sources.}
\label{app:mc11_inventory}
\end{table}

The question supplies the premise, and the default candidate hypothesis is ``The correct answer is: [option]''. GSM8K uses ``The answer is [option]'', WinoGrande substitutes the candidate into the sentence, HellaSwag uses the candidate ending, and few-shot configurations use ``Answer: [option]''. The instruction asks how this hypothesis relates to the premise, with three option descriptions: ``The hypothesis follows from the premise'', ``The hypothesis may or may not be true given the premise'', and ``The hypothesis contradicts the premise''.

Evaluation selects the first maximal candidate in the recorded order and requires three finite probabilities in $[0,1]$ whose sum differs from one by at most $10^{-5}$; an invalid output stops the evaluation. For GSM8K, the distractors perturb the correct numerical answer, and reproducing the evaluation requires the saved candidate order. Chess questions are generated with \texttt{chess} version 1.11.2 and the recorded randomization settings. Source inventories identify the dataset revisions, splits, and row order for the evaluated populations.

\subsection{Training corpus and grouped folds}
\label{app:training_data}

The predominantly English corpus contains 22,035 records and 29,897 questions drawn from web decisions, policy tasks, and public classification datasets. It combines 4,642 JevForge/Mind2Web records \citep{mind2web}, 2,676 Nimble records \citep{nimble}, 1,680 compositional-policy records, ten public-source groups of 1,000 records each, and generated policy, date, insufficient-evidence, control, and assertion records. The ten public sources are Yelp full-star reviews, DBpedia14, and AG News \citep{zhang}; Banking77 \citep{banking77}; TREC \citep{trec}; MultiNLI \citep{mnli}; IMDb \citep{imdb}; sentence-level SST-5 \citep{sst}; BoolQ \citep{boolq}; and the SetFit English packaging of MARC \citep{marc}, with all source labels converted to the shared decision schema.

JevForge uses the cleaned Mind2Web resource at revision \texttt{080a4b67} and the JevForge conversion at revision \texttt{8c6fa6c9}. The conversion preserves labels and distributions, including probability mass shared by several positive candidates. Its website partition differs from the official Mind2Web out-of-distribution split. The retained Nimble reference is revision \texttt{f136b3f75721fda4ea96\allowbreak 1f73993cc50b08488835}. Its publisher describes the reference answers as synthetic and model-checked. Training consumes record contents directly without opening external private-file or image references.

To keep related examples in the same split, we link records through shared families, paired or control examples, upstream rows, websites, and normalized states, then assign each connected group as a whole. This produces 12,472 groups, allocated with approximate feature balance: 80\% to eight training pools, 5\% to locked development, 5\% to calibration, and 10\% to a separate test split. The eight pools contain 17,661 records and 24,001 questions, with fold $j$ holding out pool $j$ and training on the remaining seven. Our four settings combine folds 2 and 3 with seeds 44 and 45: fold 2 contains 15,456 training records and requires 1,932 updates per epoch, while fold 3 contains 15,442 records and requires 1,931 updates.

The corpus contains 12,486 Choice, 12,176 binary, and 5,235 Score questions. Most supplied target distributions are one-hot (9,142 of 9,457), while the remaining distributions express insufficient evidence through 173 uniform targets or multiple valid solutions through 142 JevForge targets. Training preserves these distributions and augments compatible requests by shuffling options or adding distractor and none-of-the-above options, with augmentation probabilities ranging from 0.10 to 0.25.

\subsection{Serialization, backbone, and optimization}
\label{app:implementation}

The request format follows NeoHorse-Jev's option-conditioned implementation of typed JEV decisions~\citep{typesafe,tokenrhythm}. A record contains a state and one or more questions, each with an instruction, a type, and option descriptions or ordered levels. The encoder places the state after \texttt{<|fim\_prefix|>}, the instruction after \texttt{<|fim\_middle|>}, each option between \texttt{<|box\_start|>} and \texttt{<|box\_end|>}, and a final \texttt{<|fim\_suffix|>}. Text segments are tokenized separately without added special tokens; delimiter IDs are inserted explicitly, with no intervening spaces or newlines. The instruction precedes all option tokens. Option isolation is disabled, allowing an option to attend to earlier options in its request.

Training admits packed records of at most 4,096 tokens, with each state-plus-question sequence limited to 2,048 tokens, including delimiters. NeoHorse processes each question as a separate causal row containing the state and that question's branch. Inference permits up to 8,192 tokens per packed request and rejects longer requests. Target distributions index Choice options, false/true answers, or Score levels, assigning zero mass to missing keys before normalization. When only an integer label is supplied, it defines a one-hot target. Training minimizes $-\sum_i t_i\log p_i$ averaged over questions within a record, with the ordered-score auxiliary loss disabled.

We use \texttt{TokenRhythm/NeoHorse-1-4B} at revision \texttt{56f0584b}~\citep{neo,neocard}, with a width-256 readout and zero head dropout. Rank-64 LoRA~\citep{hu} uses $\alpha=128$ and dropout 0.05 across attention projections \texttt{q\_proj}, \texttt{k\_proj}, \texttt{v\_proj}, and \texttt{o\_proj}; gated DeltaNet projections \texttt{in\_proj\_qkv}, \texttt{in\_proj\_z}, \texttt{in\_proj\_a}, \texttt{in\_proj\_b}, and \texttt{out\_proj}; and, for all-module adaptation, feed-forward projections \texttt{gate\_proj}, \texttt{up\_proj}, and \texttt{down\_proj}. The no-FFN configuration omits these last three adapter targets while keeping the pretrained feed-forward modules.

Parameters are stored in FP32 and computation uses BF16 autocast, gradient checkpointing, and gradient clipping at 1.0. AdamW \citep{adamw} uses weight decay 0.01, microbatch size one, and accumulation over eight records. We use a cosine variant of the 1cycle policy~\citep{smith}. The schedule warms up for 10\% of updates, begins at one twenty-fifth of the peak learning rate, and cosine-anneals to one ten-thousandth of that initial rate. AdamW's first-moment coefficient $\beta_1$ cycles inversely between 0.95 and 0.85. Evaluation uses the final weights of each run. The two-epoch runs restart from saved initialization with a two-epoch schedule horizon: 3,864 updates on fold 2 and 3,862 on fold 3.

\subsection{Configurations, matching, and reuse}
\label{app:matching}

Each configuration in Tables~\ref{tab:factorial} and \ref{tab:recipe} is trained independently from matched initial parameters and uses its assigned readout throughout training and evaluation.

The pointer and mean-span heads have identical parameter counts: 1,311,232 parameters. All-module LoRA adds 129,859,584 parameters; attention and DeltaNet adapters alone add 57,507,840. The auxiliary all-module rank-28 control uses $\alpha=56$ and totals 58,124,800 trainable parameters, keeping $\alpha/r_{\mathrm{LoRA}}=2$. Its adapter shapes differ from rank 64.

Within each fold--seed setting, the core learning-rate models construct matching initial parameters using the same seed and record them before training. The adapter-scope and rank comparisons copy saved reference parameters wherever their dimensions agree and align the initial CPU/CUDA random-generator states. Separate generators control record shuffling and augmentation, with matching recorded orders, inputs, and generator traces across the comparisons. A repeated pointer executes the reference recipe again under the same recorded starting conditions.

Training the four learning-rate combinations and the repeated-pointer, trained-residual, and frozen-branch controls across four settings gives 28 models. The adapter-scope and rank comparisons add 20 models while reusing the four reference pointers; eight two-epoch models then reuse those pointers and the four all-module pooling models as their one-epoch references, giving 56 distinct trainings. The joint learning-rate experiments on H100, the additional adapter-scope comparison, and the Qwen3 experiments each use their own references.

For each named comparison, we first subtract scores within a fold--seed setting and then average the four differences. The descriptive standard error is the sample standard deviation of those differences divided by two. All arithmetic uses unrounded values; displayed means may therefore differ by 0.01 percentage points from the displayed paired difference. Factorial main effects average over both levels of the other factor, and interactions combine the corresponding four arms.

\subsection{Other endpoints}
\label{app:other_endpoints}

Our JevBench evaluation~\citep{jevbench} uses the frozen v1.2 public subset: 48 easy, 72 original, and 111 hard decisions. We report accuracy averaged equally over its 18 distinct families; this endpoint is separate from the official leaderboard composite. Invalid probability outputs count as incorrect, and ties follow the sorted option-key order. For Kev~\citep{kev}, we compute question-level accuracy in the development and test partitions of decision-v7, transfer-v4, and transfer-v9, then average the six accuracies equally. After retaining only \texttt{clean} variants and excluding the source named \texttt{unknowable}, these partitions contain 1,264/1,200, 656/656, and 1,046/1,046 questions, respectively; the test-only analysis averages the three test partitions. The fixed-checkpoint interface comparisons use all 1,838 PIQA validation questions and all 1,954 SocialIQA development questions, both outside MC11.

\subsection{Evidence coverage and reproduction}
\label{app:reproduction}

Model development used the primary evaluation suites and fold-development data. Because the four reported settings share six training pools and the same evaluation questions, their standard errors describe variation within this design. The original GPQA few-shot configuration draws demonstrations from the Diamond evaluation population; Appendix~\ref{app:additional} reports a separate evaluation with disjoint demonstrations.

Overlap screening found no exact or normalized state matches between the training corpus and 231 JevBench records or 6,436 Kev record occurrences, using Unicode NFKC normalization, case folding, and whitespace normalization. Available source identities likewise showed no overlap. Further screening covered 30,503 reranking source rows, including 285 MMLU demonstration rows, and the 198-row GPQA file. This coverage is limited to the recorded source text and identities; semantic paraphrases, substrings, pretraining exposure, synthetic chess, and rendered evaluation requests remain unassessed. The intermediate SetFit Amazon conversion has not been recovered.

The saved predictions support independent rescoring of 72 checkpoints: the 56 core NeoHorse models, eight Qwen3 models, and eight higher-rate pointer and pooling checkpoints used in the fixed-interface comparison. Candidate probabilities are stored with answer identities, candidate order, population information, and eligibility rules, allowing predictions and aggregate scores to be recomputed while preserving repeated questions and duplicate answer texts. Some additional learning-rate and adapter-scope results are available only as per-model summaries.

Reproduction requires the base weights, training records, evaluation caches, framework forks, and compatible runtimes specified by the code and configurations. The retained inputs supported reconstruction of the corpus mixture and fold exports, followed by one separate pointer/pooling pair from training through scoring on H20. The complete training corpus and numerical prediction archive are not currently available for public download. The public repository includes a small OpenCompass loader/scorer plugin, but the complete historical OpenCompass and RunWeaver forks are unavailable anonymously; redistribution permissions for RunWeaver and other assets remain unresolved. The initial source-generation and tokenizer-based eligibility steps remain unreplayed, and the tested environment retains a declared Torch dependency conflict.

CUDA determinism was disabled, allowing repeated training runs to diverge despite matched initial parameters and data streams. Dropout trajectories were not saved for the experiments that varied the head and adapter rates together. Appendix~\ref{app:residual} examines execution variation through repeated-training controls. The 56 core NeoHorse models were trained on individual GPUs of a shared eight-H20 server, using 94.46 GPU hours over the measured training intervals with allocated-memory peaks of 18.32--19.81 GiB. These measurements include concurrent-job contention and diagnostic capture; evaluation and the additional experiments fall outside the timing intervals.

%% file: supplement.tex
\section{Supplementary Comparisons}
\label{app:evidence}

\subsection{Additional training comparisons}
\label{app:cohorts}

Additional experiments test simultaneous changes to the head and adapter
learning rates, training schedules, and adapter placement. The learning-rate
and schedule comparisons use H100-trained models; the separate adapter-scope
comparison uses eight H20-trained pointers. Their lower-rate all-module
references reach 50.33\% and 49.34\% MC11, respectively, compared with
50.28\% for the core reference. Paired differences use each experiment's
own matched reference models under the same four fold--seed settings.

\subsection{Task effects, adapter count, and backbone}
\label{app:profiles}

Table~\ref{tab:supp-tasks} shows where pooling and scope restriction
help. On NeoHorse, pooling improves HellaSwag under both adapter scopes.
Removing FFN adapters from the pooling model adds 9.10~pp on HellaSwag
and 15.20~pp on chess, but lowers both GSM8K configurations and both GPQA
configurations. The corresponding pointer scope changes are also largest
on HellaSwag and chess, at 9.13 and 9.80~pp.

\begin{table}[!htbp]
\centering
\small
\setlength{\tabcolsep}{4pt}
\begin{tabularx}{\textwidth}{@{}Xrrrrr@{}}
\toprule
 & \multicolumn{4}{c}{NeoHorse} & Qwen3-4B \\
\cmidrule(lr){2-5}
Configuration & \shortstack{Pool$-$ptr.\\all} & \shortstack{Pool$-$ptr.\\no-FFN} & \shortstack{No-FFN$-$all\\pooling} & \shortstack{Pool$-$last-content\\all} & \shortstack{Pool$-$ptr.\\all} \\
\midrule
MMLU, few-shot       & $-0.34$ & $-0.38$ & $+1.32$ & $+0.94$ & $+0.16$ \\
MMLU                & $+0.49$ & $+0.46$ & $+1.42$ & $+1.75$ & $+0.01$ \\
HellaSwag           & $+4.23$ & $+4.21$ & $+9.10$ & $+8.10$ & $+3.91$ \\
GSM8K, ten options  & $-1.48$ & $-0.66$ & $-0.30$ & $-0.55$ & $-1.84$ \\
ARC-Easy            & $+1.60$ & $+0.99$ & $+1.73$ & $+4.08$ & $+0.03$ \\
GSM8K, four options & $-1.40$ & $-1.82$ & $-2.12$ & $+0.13$ & $-4.28$ \\
WinoGrande          & $+0.57$ & $+1.46$ & $+1.07$ & $+0.39$ & $-0.36$ \\
ARC-Challenge       & $+3.14$ & $+1.79$ & $+1.92$ & $+5.38$ & $+0.26$ \\
Chess               & $+1.95$ & $+7.35$ & $+15.20$ & $+0.10$ & $-1.15$ \\
GPQA, few-shot      & $+0.88$ & $-0.88$ & $-2.15$ & $+1.01$ & $-1.39$ \\
GPQA                & $-1.64$ & $-2.53$ & $-0.88$ & $-0.51$ & $-0.51$ \\
\bottomrule
\end{tabularx}
\caption{Mean within-setting task differences in percentage points.
Each column uses four paired settings; all models use one epoch and the
lower rates. The second NeoHorse column compares readouts with no-FFN
adapters; the third compares adapter scopes with pooling fixed.
The Qwen3-4B column uses a separate pointer--pooling comparison.
Numbers are computed before rounding.}
\label{tab:supp-tasks}
\end{table}

\begin{table}[!htbp]
\centering\small
\setlength{\tabcolsep}{4pt}
\begin{tabularx}{\textwidth}{@{}Xrrrr@{}}
\toprule
Kev subset & Pool$-$ptr., all & Pool$-$ptr., no-FFN & Ptr.: no-FFN$-$all & Pool: no-FFN$-$all\\
\midrule
Decision development & $+0.20\pm0.24$ & $+0.24\pm0.25$ & $-0.34\pm0.17$ & $-0.30\pm0.33$\\
Decision test & $-0.06\pm0.24$ & $-0.90\pm0.49$ & $-0.92\pm0.03$ & $-1.75\pm0.43$\\
Transfer-v4 development & $-2.44\pm0.69$ & $-2.13\pm0.65$ & $-2.71\pm1.10$ & $-2.40\pm0.27$\\
Transfer-v4 test & $-1.56\pm0.20$ & $-1.68\pm0.62$ & $-1.87\pm0.70$ & $-1.98\pm0.47$\\
Transfer-v9 development & $-3.23\pm0.53$ & $-3.30\pm0.13$ & $-2.25\pm0.51$ & $-2.32\pm0.37$\\
Transfer-v9 test & $-1.70\pm0.06$ & $-2.20\pm0.29$ & $-1.48\pm0.33$ & $-1.98\pm0.39$\\
\bottomrule
\end{tabularx}
\caption{The six Kev selections in the core readout-by-scope comparison. Entries are paired mean changes in pp with descriptive SEs. The direct-decision cost includes changes on decision-test as well as the transfer suites.}
\label{tab:supp-kev}
\end{table}

The Qwen3-4B comparison uses four lower-rate, one-epoch pairs with
rank-64 attention/FFN adapters and identical shared initial tensors.
Qwen3 has no DeltaNet targets. Pooling changes MC11 from 52.17\% to
51.70\%, a paired difference of $-0.47\pm0.59$~pp, with one positive
setting. HellaSwag retains a gain, but the ARC differences are small and
GSM8K declines. JevBench changes by $+0.17\pm1.14$~pp and Kev by
$-1.44\pm0.42$~pp. These models received no additional-source evaluation.

\subsection{Learning-rate scaling and staged adaptation}
\label{app:schedules}

The learning-rate comparison lowers the head and adapter rates together while
retaining matched initial tensors and the one-epoch schedule. Pointer
MC11 increases from 45.90\% to 50.33\%; the residual-head and mean-span pooling models
increase by 1.60 and 1.75~pp. Consequently, their advantages over the
same-rate pointer shrink from 3.41 to 0.57~pp and from 3.95 to
1.26~pp. Here both rates change together; the adapter-rate main effect in
Table~\ref{tab:recipe} averages over two head rates.

Head-first adaptation freezes the backbone for the first
$\lceil0.3N\rceil$ updates and then enables adapters without restarting
the full-horizon schedule. The backbone runs in evaluation mode during
the initial phase. Head-only-tail training instead places the same
number of head-only updates at the end. Both schedules omit 580 adapter
updates per setting but expose the adapters to different examples,
learning rates, momentum histories, and dropout phases.

\begin{table}[!htbp]
\centering
\small
\begin{tabularx}{\textwidth}{@{}Xrrr@{}}
\toprule
Pointer recipe or paired change & MC11 & JevBench & Kev \\
\midrule
Joint training, higher rate (accuracy, \%) & 45.90 & 75.62 & 81.47 \\
Lower-rate joint $-$ higher-rate joint (pp) & $+4.43\pm1.03$ & $+2.35\pm0.67$ & $-0.43\pm0.53$ \\
Head-first $-$ higher-rate joint (pp) & $+6.38\pm2.17$ & $+2.03\pm0.84$ & $-1.02\pm0.28$ \\
No-FFN $-$ higher-rate joint (pp) & $+3.67\pm1.33$ & $+1.80\pm1.53$ & $-0.02\pm0.34$ \\
Head-first + no-FFN $-$ higher-rate joint (pp) & $+7.64\pm1.91$ & $+0.96\pm0.66$ & $-1.95\pm0.40$ \\
Head-only tail $-$ higher-rate joint (pp) & $+3.84\pm0.88$ & $-0.31\pm1.40$ & $-0.30\pm0.23$ \\
\bottomrule
\end{tabularx}
\caption{Effects of update timing and adapter placement in a
separately trained set of pointers. All recipes use the option-ending pointer. Except for the
explicit lower-rate row, both peak rates are $2.5\times10^{-5}$.
Differences are means $\pm$ descriptive SE across four settings.
Head initialization can vary when adapter scope changes in these comparisons.}
\label{tab:supp-schedules}
\end{table}

Head-first reaches 52.28\% MC11 and exceeds head-only tail by
$2.54\pm1.49$~pp. Combining head-first and no-FFN reaches 53.54\%,
but its scope-by-schedule interaction is $-2.41\pm1.61$~pp: the
combined gain is smaller than the sum of the separate gains.
The combined head-first/no-FFN model uses the higher-rate pointer recipe;
adding pooling to this recipe remains untested. The separate adapter-scope
comparison starts from identical head and attention/DeltaNet adapter tensors.

A further seven-pair retraining checks a different rate interval,
$5\times10^{-5}$ to $2.5\times10^{-5}$ for both parameter groups.
The mean MC11 change is $-0.34$~pp: five pairs improve, while two
decline by 2.46 and 6.10~pp. JevBench and Kev means rise by 1.69 and
0.74~pp. Those pairs share their seeded construction procedure, but
saved initial-tensor identity is not established. The rate result
therefore depends on the interval and comparison being tested.

\subsection{Additional sources and inference execution}
\label{app:additional}

\paragraph{Fixed-question requests.}
The PIQA state is ``Goal: [goal]'', followed by a newline and ``Which solution best achieves this goal?''. Direct selection presents both solutions under the instruction ``Select the solution that best achieves the goal.'', whereas reranking evaluates each solution in a separate NLI request with the hypothesis ``The correct solution is: [solution]''. For the same 1,838 questions, these interfaces require one and two requests per question, respectively.

SocialIQA version 1.4 combines the original context, two newlines, and the question as its state. Direct selection presents the raw A/B/C answer texts under ``Select the correct answer to the question.''; NLI reranking places each raw answer in quotation marks within ``Hypothesis: [raw answer] How does it relate to the premise?''. Answer fragments remain in their original form, and the three NLI options use the descriptions in Appendix~\ref{app:mc11_protocol}. Comparing entailment probabilities across answers requires three requests per question, against one for direct selection. All 1,954 questions are evaluated with eight one-epoch checkpoints from the core all-module comparison and eight two-epoch checkpoints from the budget comparison.

The SocialIQA source, checkpoint selection, candidate order, request templates, and overlap screening were specified before scoring, after the main evaluations had begun. Exact matching, normalized lexical containment, and available source-lineage checks found no overlap with the training inputs and derivation files, so all questions were retained. These checks address recorded text and provenance; semantic overlap and pretraining exposure remain unmeasured.

\paragraph{NLI descriptions and option order.}
To examine sensitivity to the wording of the NLI options, we use the default descriptions in Appendix~\ref{app:mc11_protocol} and two alternatives. The second set is ``The premise supports the hypothesis as true.'', ``The premise does not determine whether the hypothesis is true.'', and ``The premise supports the hypothesis as false.'' The third is ``The hypothesis must be true if the premise is true.'', ``The hypothesis could be either true or false if the premise is true.'', and ``The hypothesis must be false if the premise is true.'' For each fixed checkpoint, we cross these descriptions with all six label orders, giving 18 templates while keeping external solution order fixed and retrieving entailment by label identity.

\paragraph{Additional-source protocol.}

PIQA and CommonsenseQA evaluate 1,838 validation and 1,221 development
questions through NLI reranking. CLINC150 \citep{clinc} evaluates all 4,500 in-scope
test utterances through a single 150-option direct request. These
evaluations cover 36 models varying learning rates, training schedules,
or adapter placement, plus eight separately trained pointers comparing
adapter scopes with matched initialization. The final no-FFN pooling
configuration is evaluated on MC11, JevBench, and Kev.

CommonsenseQA uses the question stem without its question-concept field;
each of its five candidate answers becomes ``The correct answer is: [candidate]''.
The 24 questions with duplicate answer texts retain distinct candidate identities.
Reranking candidate order follows the SHA-256 of canonical JSON triples
containing the source revision, sample identity, and candidate identity,
with candidate identity as a secondary key. Gold labels enter neither
requests nor ordering. Exact ties use the first maximum. Uniform-tie credit
leaves the reported pooling and scope effects unchanged. Registered training
and development-selection materials contain no exact or normalized PIQA or
CommonsenseQA question matches under Unicode NFKC, case, and whitespace
normalization; semantic overlap and pretraining exposure remain unmeasured.

On CLINC150, pooling lowers accuracy by $3.64\pm0.54$~pp in the
lower-rate readout comparison, and omitting FFN adapters lowers it by
$1.78\pm0.41$~pp in the separate scope comparison. These direct-selection
losses accompany the PIQA and CommonsenseQA gains in Table~\ref{tab:extra}.
The known CLINC overlap item is retained and correctly classified by every
model, contributing zero to each paired difference.

The fixed-checkpoint PIQA and additional-source comparisons evaluate the
same eight higher-rate models in two execution environments. The original
inference service on H100 gives pointer/pooling accuracies of 65.27/73.12\%,
while the packaged-model service on H20 gives 65.15/73.22\%. These close
aggregate scores conceal changes in about 2.6\% of the 14,704 model--question
selections, with nearly balanced correct-to-wrong and wrong-to-correct
changes. We therefore keep each comparison within one environment:
the interface tests use the packaged-model service, and the additional-source
tests use the original service. The two executions match weights, tokenizer,
model and loading code, requests, labels, and candidate order; both use a
BF16 backbone and FP32 head in evaluation mode. Hardware and service
environment vary together.

Wording changes the readout comparison even with the checkpoints fixed.
Across the 18 templates, the mean pooling effect is $+3.18\pm3.27$~pp,
but averaging label orders within each description gives +6.31,
$-1.80$, and +5.03~pp, and the smallest fixed-template mean is $-5.51$~pp.
The sign changes show why the default-template gain should be interpreted
alongside this sensitivity to how the NLI relations are described.

For GPQA, a separate five-shot evaluation excludes Diamond identities,
normalized question matches, and revision aliases from the Main-split
demonstration pool. On the same 198 Diamond questions and eight
higher-rate checkpoints, pooling scores 36.87\% against the pointer's
33.46\%, a $+3.41\pm2.75$~pp effect with two positive settings.
This additional evaluation changes demonstration selection, candidate order,
and identity handling together. Its scores are reported separately, with
MC11 retaining the original GPQA configurations.

\section{Residual Corrections and Training Controls}
\label{app:residual}

The residual head augments the option-ending pointer with a learned
decision--option correction. For projected query and key vectors
$q,k_i\in\mathbb{R}^{256}$ and the bilinear pointer score $s_i$, it uses
\begin{align}
v_i &= [q;k_i;q\odot k_i;|q-k_i|], \nonumber\\
s_i^{\mathrm{pair}} &= s_i+w^\top\operatorname{GELU}(Av_i+a)+b.
\label{eq:residual-app}
\end{align}
The correction adds a 128-unit hidden layer and 131,329 parameters, with
output weights $w$ and bias $b$ initialized to zero. To examine its role
during training, we compare optimization of all four branch tensors with
a frozen branch whose output remains zero while the shared projections
and adapters train. To examine its contribution after training, we remove
the correction from a trained residual model and apply the softmax to $s_i$.

\subsection{Removing the correction from a trained model}

For each of the four higher-rate residual models, we compare corrected
and pre-correction logits from the same forward pass, keeping the trained
backbone and query/key projections fixed. The head computes both sets of
logits in FP32 without autocast, so the comparison isolates the correction's
contribution at inference within an already trained model.

Removing the correction changes only four of 185,900 selected answers,
with no correct answer becoming incorrect, and increases mean MC11 by
less than 0.001 pp. This mean uses the same configuration weighting as
the main evaluation. The learned bilinear component therefore recovers
virtually all answer selections at these checkpoints.

\subsection{Frozen branches, repeated training, and divergence}

The training comparisons show a stronger dependence on the recipe.
At the higher rate, the frozen-branch model exceeds the pointer by
$1.39\pm0.38$~pp on MC11, and the trained residual model exceeds the
frozen-branch model by a further $2.02\pm0.97$~pp. Under the lower-rate recipe, the core comparisons are
small and depend on the reference run (Table~\ref{tab:supp-residual}),
while the separate comparison with jointly reduced head and adapter rates
retains a $+0.57\pm0.18$~pp residual advantage.

\begin{table}[!htbp]
\centering
\small
\begin{tabularx}{\textwidth}{@{}Xrrr@{}}
\toprule
Lower-rate training comparison & MC11 & JevBench & Kev \\
\midrule
Repeated pointer $-$ reference & $+0.17\pm0.74$ & $-0.66\pm0.52$ & $-0.15\pm0.37$ \\
Frozen branch $-$ reference & $+0.40\pm0.93$ & $-0.13\pm1.60$ & $-0.06\pm0.52$ \\
Trained residual $-$ reference & $+0.05\pm0.57$ & $-0.82\pm1.61$ & $-0.11\pm0.41$ \\
Trained residual $-$ frozen branch & $-0.35\pm0.59$ & $-0.69\pm0.87$ & $-0.04\pm0.22$ \\
Trained residual $-$ repeated pointer & $-0.12\pm0.58$ & $-0.16\pm1.82$ & $+0.04\pm0.20$ \\
\bottomrule
\end{tabularx}
\caption{Lower-rate residual controls, in percentage points (mean
$\pm$ descriptive SE over four settings). All arms use one epoch and
all-module rank-64 adapters. The repeated pointer reruns the reference
recipe with the same nominal seed.}
\label{tab:supp-residual}
\end{table}

Repeating the pointer from the same recorded initial parameters, nominal
seed, and recorded data streams changes its MC11 score enough to reverse
the small residual-versus-pointer contrast (Table~\ref{tab:supp-residual}).
Short matched probes help explain why identical starting conditions are
insufficient: even pairs of ordinary pointers develop different gradients
and parameters when CUDA determinism is disabled. The operations responsible
and their contribution to the full-training differences remain unresolved.

During lower-rate residual training, corrected and pre-correction BF16
logits select the same option on all recorded training questions.
BF16 addition can round away a nonzero correction. These traces concern
option choices during lower-rate training, whereas the FP32 intervention
measures reranking selections at higher-rate checkpoints.

\FloatBarrier
\section{Aggregation and Reference Sensitivity}
\label{app:robust}

MC11 gives equal weight to eleven configurations. The seven-source
alternative first averages configurations within ARC, GSM8K, GPQA,
and MMLU, then gives these four means and HellaSwag, WinoGrande, and
chess equal weight. Configuration-weighted leave-one-source-out (LOSO)
removes all configurations of one source and averages the remainder;
source-weighted LOSO averages the remaining six source means. Every
omission retains all four model settings. The ranges in
Table~\ref{tab:supp-weights} summarize how the mean effects change with
benchmark composition.

\begin{table}[!htbp]
\centering
\small
\begin{tabularx}{\textwidth}{@{}Xrrrr@{}}
\toprule
Contrast & MC11 & \shortstack{Seven\\sources} & \shortstack{Config.-weighted\\LOSO} & \shortstack{Source-weighted\\LOSO} \\
\midrule
Head-rate main effect & $+0.02\pm0.43$ & $-0.11\pm0.46$ & $[-0.132,+0.285]$ & $[-0.345,+0.314]$ \\
Adapter-rate main effect & $-2.83\pm0.79$ & $-2.96\pm0.65$ & $[-3.393,-2.240]$ & $[-3.408,-2.478]$ \\
Pooling $-$ pointer, all & $+0.73\pm1.07$ & $+1.05\pm1.11$ & $[+0.363,+1.209]$ & $[+0.524,+1.470]$ \\
Pooling $-$ pointer, no-FFN & $+0.91\pm0.35$ & $+1.64\pm0.57$ & $[+0.264,+1.489]$ & $[+0.693,+2.202]$ \\
No-FFN $-$ all, pointer & $+2.21\pm0.66$ & $+3.10\pm0.64$ & $[+1.451,+3.015]$ & $[+1.985,+3.854]$ \\
No-FFN $-$ all, pooling & $+2.39\pm0.35$ & $+3.69\pm0.40$ & $[+1.110,+3.259]$ & $[+1.773,+4.558]$ \\
\mbox{Readout-by-scope} interaction & $+0.18\pm0.74$ & $+0.59\pm0.66$ & $[-0.341,+0.515]$ & $[-0.212,+0.909]$ \\
No-FFN rank 64 $-$ all rank 28 & $+0.09\pm0.42$ & $+0.51\pm0.59$ & $[-0.200,+0.514]$ & $[+0.102,+0.900]$ \\
Mean-span $-$ last-content & $+1.89\pm0.97$ & $+2.10\pm0.77$ & $[+1.265,+2.362]$ & $[+1.103,+2.488]$ \\
Trained residual $-$ reference & $+0.05\pm0.57$ & $-0.09\pm0.64$ & $[-0.343,+0.338]$ & $[-0.407,+0.366]$ \\
Budget interaction & $+3.88\pm1.13$ & $+3.66\pm1.37$ & $[+2.530,+4.553]$ & $[+2.608,+4.428]$ \\
\bottomrule
\end{tabularx}
\caption{Sensitivity of learning-rate, readout, adapter-placement,
and training-budget effects to benchmark aggregation (pp). The first two columns are paired means
$\pm$ descriptive SE. Each LOSO range spans seven omission-specific
means. Main effects use higher minus lower rate; the budget interaction
is the two-epoch pooling advantage minus the one-epoch advantage.}
\label{tab:supp-weights}
\end{table}

The head-by-adapter rate interaction is $+1.28\pm1.65$~pp on MC11.
The pooling and no-FFN main comparisons retain positive four-setting means
under both weighting rules and every source omission. The interaction
between them is less stable: omitting chess reverses its MC11-weighted
mean. Likewise, no-FFN rank~64 versus all-module rank~28 can change sign
under configuration-weighted omission. The near-zero head-rate and
trained-residual means change sign under equal source weighting.
These outcomes support the individual recipe comparisons more strongly
than claims about their interactions or the ranking of near-equal arms.

The budget result also persists under reweighting: longer training lowers
both models' accuracy, with the pointer losing more. ARC and HellaSwag
account for about 63\% of the wider gap. Under seven-source weighting,
the mean interaction remains +3.66~pp, although one setting has a small
negative interaction of $-0.20$~pp.

Replacing the reference pointer by its repeated training changes the
all-module pooling effect to $+0.55\pm1.11$~pp and the trained-residual
effect to $-0.12\pm0.58$~pp. No-FFN and rank-28 pointers retain gains
of $+2.04\pm0.63$ and $+1.95\pm0.77$~pp, each positive in all four
settings. These larger recipe gains are less dependent on which
reference execution is used.

Pooling and scope restriction also lower Kev under the three-test-subset
average, with declines on the decision test as well as transfer subsets.
The smaller residual-control contrasts change sign under this weighting.